\documentclass[runningheads]{llncs}

\usepackage{eccv}

\usepackage{eccvabbrv}

\usepackage{graphicx}
\usepackage{booktabs}
\usepackage{wrapfig,lipsum}
\usepackage[accsupp]{axessibility}  

\usepackage{makecell}
\usepackage{pifont,xcolor}
\usepackage{siunitx} 
\usepackage{multirow}
\usepackage{colortbl}
\usepackage{wrapfig}
\newcommand{\cmark}{\textcolor{ForestGreen}{\ding{51}}} 
\newcommand{\xmark}{\textcolor{red}{\ding{55}}}    
\usepackage{subcaption}
\newcommand{\NA}{\textcolor{lightgray}{N/A}}

\usepackage{float}
\usepackage{caption}
\usepackage{bibunits}
\usepackage{setspace}

\definecolor{extralightgray}{gray}{0.90}

\usepackage{hyperref}

\usepackage{orcidlink}

\begin{document}
\begin{bibunit}[splncs04]

\title{
Localizing to Debias: A Patch-Level Benchmark and Baseline for Weakly Supervised Spatial Anomaly Detection
} 
\titlerunning{A Benchmark and Baseline for WSVAD Localization}


\author{Sara Abdulaziz\inst{1}\orcidlink{0009-0002-5204-127X} \and
Abdulrahman Al-Abri\inst{1}\orcidlink{ } \and
Giacomo D'Amicantonio \inst{1}\orcidlink{0000-0002-3783-6464} \and Egor Bondarev \inst{1}\orcidlink{0009-0005-2452-7389} }

\authorrunning{S.~Abdulaziz et al.}

\institute{Eindhoven University of Technology, 5612 AE Eindhoven, Netherlands\\ \email{s.e.a.m.abdulaziz@tue.nl}\\
\href{https://github.com/Sara-Esam/SST-WSVADL}{GitHub Code}
}

\maketitle

\begin{abstract}

  Despite growing interest in weakly supervised video anomaly detection (WSVAD), current methods struggle to bridge the gap between coarse temporal supervision and fine-grained spatial reasoning. A key obstacle is the tendency of temporal detectors to latch onto background and scene-level cues rather than truly discriminative anomaly evidence. This background bias raises ethical concerns: models may inadvertently associate anomalies with societal or environmental context rather than authentic crime-related cues. Without spatial grounding, such biases remain hidden and unauditable. To address this, we propose SST-WSVADL, a sparse spatio-temporal framework that bridges temporal anomaly detection with fine-grained spatial localization. Rather than processing all spatial regions indiscriminately, SST-WSVADL progressively focuses on the most anomaly-relevant spatio-temporal regions through dynamic sparsification, naturally suppressing background dominant content while preserving discriminative evidence. The temporal and spatial branches are coupled end-to-end via motion-aware regularization that guides sparsification toward dynamically informative regions, without relying on external detectors or vision-language prompts. We publicly release frame-level spatial annotations and a method-agnostic evaluation protocol for three public datasets: UCF-Crime, XD-Violence, and MSAD. These resources enable the community to audit spatial biases in WSVAD predictions, supporting progress toward more ethical and accountable anomaly detection. Experiments demonstrate that SST-WSVADL is competitive with prior methods across benchmarks while enabling localization and patch-level auditability of scene bias, providing a reproducible foundation for interpretability-oriented evaluation of WSVAD models.


\end{abstract}

\section{Introduction}
\label{sec:intro}

Video Anomaly Detection (VAD) has been a long-standing challenge due to the rarity, diversity, and context dependence of anomalous events. In practice, the cost and ambiguity of annotation make fully supervised spatio-temporal labeling impractical at scale. Consequently, weakly supervised VAD (WSVAD) has emerged as the dominant paradigm, where models are trained using only video-level labels while inferring fine-grained temporal and spatial anomalies at test time.  However, this weak supervision induces a structural limitation: rich spatio-temporal evidence is compressed into coarse segment-level scores, often causing models to rely on global scene context rather than truly discriminative local motion cues. Recent studies further reveal that such global representations amplify background bias \cite{liu2019exploring}, where static scene elements dominate anomaly predictions and degrade localization reliability. Systematically addressing this transparency concern remains an open challenge in WSVADL. Beyond performance, this bias carries ethical implications: models that associate anomalies with background scene context, such as location or ethnic attributes, rather than actual criminal behavior risk making decisions that are unfair to certain societal segments. For instance, flagging ordinary cultural practices or daily routines as suspicious simply because they occur in low-income or minority-dense neighborhoods that are overrepresented in training data for crimes like robbery or assault.


To address these limitations, recent efforts have extended WSVAD toward weakly-supervised spatio-temporal anomaly detection and localization (WSVADL), yet existing approaches consistently trade one external dependency for another, e.g. object detectors~ \cite{wu2021weakly}, dense annotations~\cite{liu2019exploring}, or vision-language models~\cite{stprompt_wu2024weakly}, to guide localization. Crucially, none of these strategies tackle background bias end-to-end, leaving interpretability and domain robustness as persistent open concerns. These limitations are further compounded by the lack of spatial annotations and unified evaluation protocols: existing works report a mix of frame-AUC, temporal-IoU~\cite{liu2019exploring,stprompt_wu2024weakly}, mean-IoU \cite{wu2021weakly}, or prompt-based heatmaps\cite{sun2023long}, preventing consistent and fair comparison across methods.

In this work, we address both challenges jointly. We curate and release spatial anomaly annotations for three public datasets: UCF-Crime, XD-Violence, and MSAD, paired with a method-agnostic patch-level evaluation protocol that enables fair, reproducible, and bias-auditable comparison. Building on this foundation, we introduce SST-WSVADL, a lightweight end-to-end framework that grounds anomaly evidence spatially through structured sparsification, without relying on external detectors, dense annotations, or vision-language models. In short, the contributions of this work are three-fold:

\begin{enumerate}
    \item We release a unified spatial anomaly localization benchmark for UCF-Crime, XD-Violence, and MSAD, together with a method-agnostic patch-level evaluation protocol that enables reproducible comparison and auditing of ethical biases in WSVAD predictions. 
    
    \item We propose SST-WSVADL, a \textit{detector-free, annotation-light, and non-VLM} framework that unifies snippet-level temporal reasoning with motion-aware patch-level localization under weak supervision, without external dependencies.
    
    \item We provide a transparent baseline for the unified benchmark, along with a reproducible and interpretability-oriented evaluation of spatial bias in WSVADL methods.
\end{enumerate}

\section{Related Work}

Video anomaly detection (VAD) aims to identify events that deviate from normal behavioral patterns in surveillance videos. Traditional approaches relied on handcrafted features and shallow models, such as sparse representation \cite{lu2013abnormal,zhao2011online} or motion-based descriptors \cite{cui2011abnormal, hospedales2009markov,kratz2009anomaly, li2013anomaly,zhu2012context}, to detect deviations from learned regularity motion patterns. With the success of deep learning, Sultani et al. \cite{sultani2018real} introduced the UCF-Crime dataset and the first deep multiple instance learning (MIL) framework, which became the foundation for most weakly-supervised video anomaly detection (WSVAD) methods.

Early studies on WSVAD treated videos as globally averaged features, processing full-frame representations for anomaly detection, while overlooking the local spatial variation and motion cues \cite{sultani2018real}. Landi et al. \cite{landi2019anomaly} demonstrated that anomalies are inherently local, and occupy a limited region of space and time, and proposed analyzing spatio-temporal tubes instead of entire frames. Their work established UCFCrime2Local, the first dataset with bounding-box annotations, showing that focusing on localized regions improves detection accuracy and robustness. Complementarily, Liu and Ma \cite{liu2019exploring} revealed the background bias problem, where global representations inadvertently correlate with static context rather than actual motion or actions, thereby misguiding anomaly detection models. The authors proposed a supervised anomaly localization framework that later motivated spatially-focused and motion-aware designs \cite{wu2021weakly,li2022scale,sun2023long}. Wu et al. \cite{wu2021weakly} introduced WSSTAD, the first framework to jointly reason over space and time under weak supervision. It models anomalies at the tube level using spatio-temporal proposals instead of frame averages, bridging the gap between anomaly detection and action localization. Similarly, Li et al. \cite{li2022scale} proposed a scale-aware spatio-temporal relation learning approach that captures dependencies across multiple granularities using transformers. Sun et al. \cite{sun2023long} advanced temporal modeling through long-short temporal co-teaching, improving snippet consistency and label reliability in weak supervision settings. More recently, vision-language models (VLMs) have reshaped WSVAD by introducing semantic priors through text prompts. Wu et al. \cite{stprompt_wu2024weakly} proposed STPrompt, which leverages spatio-temporal prompting and large language models (LLMs) for training-free spatial localization, unifying temporal detection and spatial reasoning. While this method demonstrated promising results, its dependence on large pre-trained VLMs and limited-context models, such as CLIP, constrains both flexibility and interpretability. In contrast, our framework is \textit{detector-free, non-VLM, and annotation-light}, leveraging \textit{learnable sparsity} for efficient and fully end-to-end anomaly localization.


\section{Datasets and Spatial Annotations} \label{sec:dataset}
We build upon previous efforts on anomaly localization~\cite{landi2019anomaly, liu2019exploring} and extend spatial annotations to full UCF-Crime, XD-Violence, and MSAD datasets. Whereas a \textit{temporal} annotation captures \textbf{when} an anomaly occurs, a \textit{spatial} annotation captures \textbf{where} it occurs via per-frame bounding boxes. The anomaly box encloses all actors and relevant objects involved in the event. The temporal extent of an anomaly begins when all required actors are visibly engaged in the anomalous action and ends when they leave the scene or the action ends. The set of frame-wise boxes over this interval constitutes a video tube, which forms the annotation. Our annotation protocol is designed to produce temporally stable, semantically consistent bounding boxes that align with the patch-level evaluation granularity. It proceeds in two stages:



\begin{enumerate}
    \item\textbf{Event timing.} Anomalous events are temporally bounded by identifying onset and offset frames, distinguishing between anticipatory context and the anomaly itself.

    \item\textbf{Spatial localization.} For each event, rather than annotating tight per-frame boxes, we adopt a semantically-driven update policy that reflects the underlying event structure:
    \begin{itemize}
        \item \emph{Stationary events.} A single box is maintained for the event duration, anchored to the region of occurrence rather than individual actor positions.
        \item \emph{Non-stationary events.} The box is updated only when the event has shifted substantially beyond the current boundary, prioritizing semantic continuity over frame-level precision.
    \end{itemize}
\end{enumerate}

This procedure differs from prior annotation efforts on UCF-Crime~\cite{landi2019anomaly, liu2019exploring}, which produce excessively shifting boxes that introduce noise and reduce evaluation consistency. By contrast, our semantically-driven update policy yields temporally stable annotations that better reflect the true spatial extent of anomalous events, as demonstrated in Supp. S1.1. Detailed annotation statistics including bbox-to-frame ratios, tubelet density, and per-category breakdowns are reported in the Supp. S1.2.

\section{Sparse Spatio-Temporal WSVADL Framework}

\begin{figure*}[htb!]
    \centering
    \includegraphics[width=0.9\linewidth]{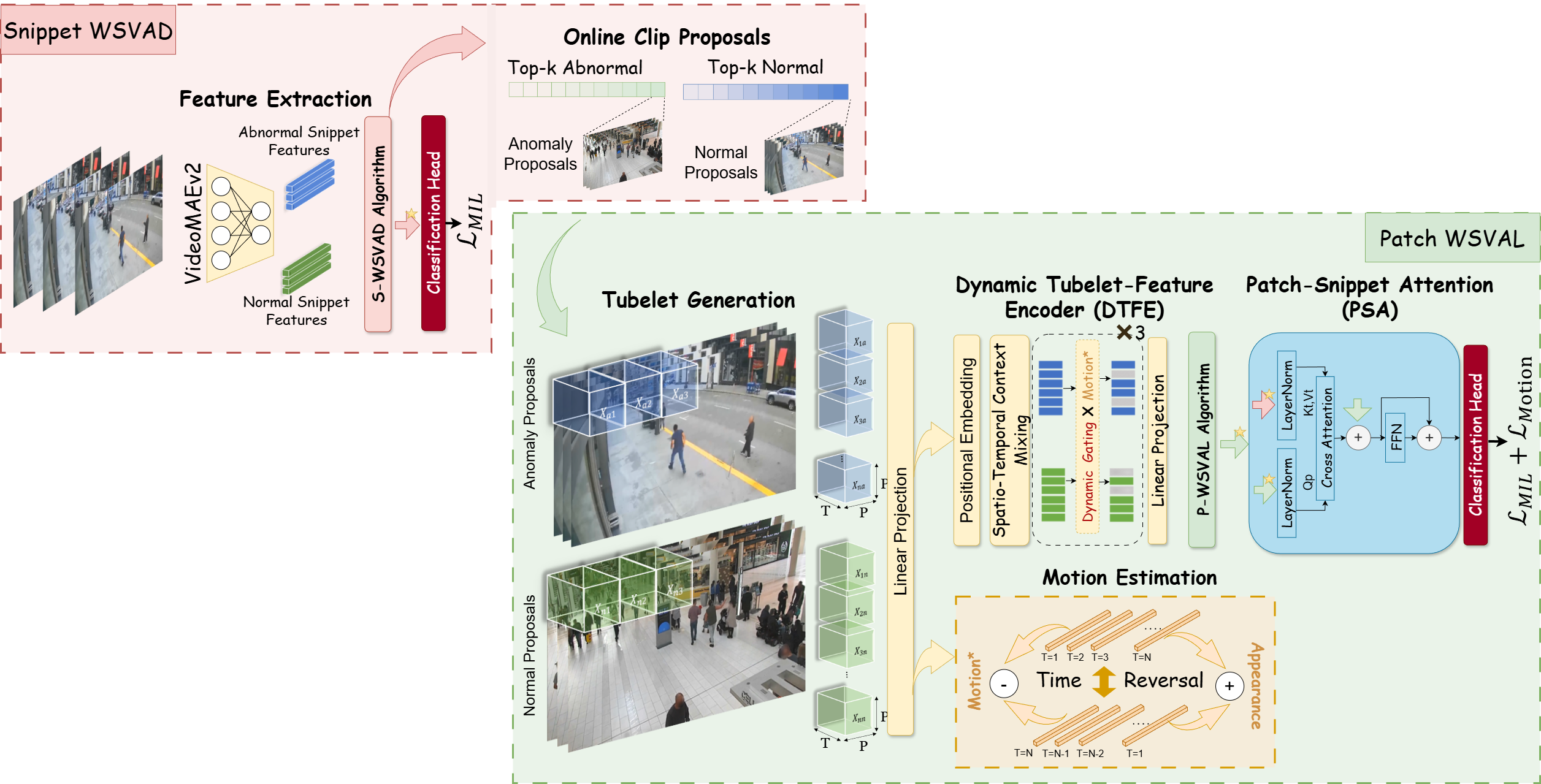}
    \caption{Overview of the proposed SST-WSVADL. The framework consists of two interacting branches. (Left) The Snippet-WSVAD branch is a standard WSVAD model \cite{urdmu_zhou2023dual} that processes snippet-level features extracted via a pre-trained task-agnostic backbone \cite{videomaev2}. The normal and abnormal clips for the patch branch are selected through the Online Clip Proposal (OCP) module. (Right) The Patch-WSVAL branch converts snippet proposals into tubelets, encodes them by the Dynamic Tubelet-Feature Encoder (DTFE), and computes motion scores that bias DTFE towards dynamic tubelets. A Patch-Snippet Attention (PSA) module fuses the compact spatial and temporal features before the final classification head. }
    \label{fig:framework}
\end{figure*}

The proposed Spatio-Temporal framework, SST-WSVADL, integrates two complementary branches. The \textbf{snippet-level branch (S-WSVAD)}, which performs temporal anomaly detection over video snippets, and the \textbf{patch-level branch (P-WSVAL)}, which localizes anomalies spatially based on sparsified tubelets. The snippet branch provides temporal anomaly scores and generates clip proposals for the patch branch. The latter generates sparse tubelet representations for localization and provides a feedback to refine the temporal detection. Figure \ref{fig:framework} illustrates the overall framework.

\subsection{Snippet-level WSVAD} 
The snippet branch follows a standard MIL-based WSVAD approach \cite{sultani2018real}, processing input videos as sequences of fixed-length snippets. Features are extracted with a pre-trained backbone and passed through a WSVAD algorithm \cite{urdmu_zhou2023dual}. We choose task-agnostic features for this branch, produced from a self-supervised VideoMAEv2~\cite{videomaev2} trained on large scale video data and fine-tuned for action recognition. This enables better generalization, and finer motion-appearance disentanglement than the commonly adopted I3D backbone \cite{i3d}. 


\paragraph{Online Clip Proposal Module.}

The online clip proposal module leverages the ability of the chosen WSVAD algorithm \cite{urdmu_zhou2023dual} to select the most discriminative video segments for the patch branch training. The algorithm stores prototypes of normal and abnormal samples in separate learnable memory banks. During training, the algorithm applies self-attention between the snippet features and the memory banks to assess the similarity of features to the stored prototypes. Then, the top-k features are selected based on the highest similarity to the prototypes. We leverage this property of scanning stored prototypes to propose the most representative video snippets for patch branch training. Concretely, we select the first from the top-k normal and abnormal features to train the patch branch. This proposal mechanism ensures the patch branch is trained on hard discriminative regions, rather than processing all video segments indiscriminately. In addition, since the exact temporal evidence provided by the MIL loss in the temporal branch is applied to the patch branch, the two branches exchange information about the same features, thereby guaranteeing alignment of their predictions.








\subsection{Patch-level WSVAD} \label{sec:patch-wsvad}
Complementing the snippet-level reasoning, the patch branch focuses on spatial anomaly localization within each frame. However, processing all patches exhaustively is computationally demanding, and more critically, risks diluting anomaly evidence with overwhelming background content, reinforcing the bias we aim to suppress. To address this, we localize anomalies by progressively focusing on the most relevant spatio-temporal regions. The Online Clip Proposal (OCP) module first narrows the candidate clips to those aligned with anomaly prototypes. Subsequently, dynamic sparsification adaptively discards VAD-irrelevant tubelets while retaining those critical for localization.

The patch-level branch consists of four components: a tubelet generator, the Dynamic Tubelet Feature Encoder (DTFE), Patch-Snippet Attention (PSA), and a WSVAD objective. DTFE integrates local, spatial, and spatio-temporal cues while progressively pruning uninformative tubelets, producing a compact set of discriminative representations. These retained tubelets are fused with snippet-level features through PSA and optimized for spatial anomaly localization. It is worth noting that the primary role of dynamic sparsification is not computational reduction only, but to enable structured feature selection under weak supervision. By suppressing background-dominant tokens, it prevents spatial reasoning from defaulting to global scene context and grounds localization in discriminative anomaly cues.

\paragraph{Tubelet Generation.}
A video segment of $F$ frames is first divided temporally into $t_z$ sub-segments of equal duration. Within each sub-segment, every frame is partitioned into a spatial grid of non-overlapping patches of size $P \times P$ pixels, yielding $T = \tfrac{H}{P} \times \tfrac{W}{P}$ spatial locations per sub-segment. Patches at the same spatial location across the frames of a sub-segment are stacked to form a \emph{tubelet}, giving $N = t_z \times T$ tubelets per segment in total. The split size $t_z$ controls the temporal resolution. 


\paragraph{Dynamic Tubelet-Feature Encoder (DTFE).}
For tubelets encoding, we adopt a tubelet-based Transformer design that follows a dynamic token sparsification strategy \cite{rao2021dynamicvit, li2023stprivacy}. The full mathematical formulation is provided in Supp. 2.1.

\paragraph{Patch-Snippet Attention (PSA).}  

To further align the two branches, we introduce a \emph{patch-snippet context attention} module. Let $S = \{s_t\}_{t=1}^T$ denote the final snippet-level features from the temporal branch (S-WSVAD), and $Y = \{y_i\}_{i=1}^{N'}$ the spatio-temporal token embeddings from DTFE. We model their interaction by a cross-attention mechanism, where patch tokens attend to the snippet-level context  

\begin{equation}
    \mathrm{Attn}(Y, S) = 
    \mathrm{Softmax}\!\left(\frac{Q(y_i)K(S)^\top}{\sqrt{d}}\right)V(S),
\end{equation}  
with $Q, K, V$ being learnable linear projections. The updated patch representation is given by  

\begin{equation}
    \hat{y}_i = y_i + \mathrm{Attn}(y_i, S).
\end{equation}  

This context mixing ensures two properties: (1) patch-level anomaly localization remains temporally consistent with snippet predictions, and (2) gradients from patch-level branch flow back into snippet-level features, thereby providing stronger supervision for the temporal branch.


\subsection{Training Procedure} 

The two-branch framework is trained end-to-end. In line with the standard MIL-based WSVAD approaches \cite{sultani2018real, urdmu_zhou2023dual}, we form snippet pairs from normal and abnormal sets for the S-WSVAD branch, which later produces proposals by the OCP module. These proposals are processed by P-WSVAL to generate normal and abnormal tubelets for localization. Following prior optimization strategies \cite{urdmu_zhou2023dual}, we train both snippet and patch-WSVAD branches with a binary cross-entropy (BCE) loss $L_{cls}$ added to four other supplementary losses $L_{dm}$, $L_{trip}$, $L_{kl}$, $L_{dis}$ \cite{urdmu_zhou2023dual}


\begin{equation}
\mathcal{L}_{WSVAD} = L_{\text{cls}} 
+ \lambda_1 L_{dm} 
+ \lambda_2 L_{trip} 
+ \lambda_3 L_{kl} 
+ \lambda_4 L_{dis},
\end{equation}
where $\lambda_1,\lambda_2,\lambda_3,\lambda_4$ are hyperparameters. 

\subsubsection{Motion Regularization} In P-WSVAL, we introduce a motion-aware regularization term to mitigate background shortcut learning under weak supervision. Without additional bias, dynamic token sparsification may retain visually salient yet motion-irrelevant regions (static context), leading to unstable spatial reasoning. We impose a motion prior that encourages DTFE to preserve dynamically informative tubelets during pruning. Instead of extracting motion at the RGB level (e.g., optical flow), which is computationally expensive at the tubelet scale, we model motion directly in the embedding space. 




Building on principles from spatio-temporal representation learning \cite{wang2019self, wang2021tdn} and symmetry decomposition~\cite{reynold}, we split the tubelet embedding into \emph{appearance} (time-symmetric) and \emph{motion} (time-antisymmetric) components, following the intuition that motion cues change under temporal reversal while appearance cues do not. This allows isolating motion through extracting the time-antisymmetric component from the tubelet embedding. For each tubelet $t_i$, we form its time-reversed counterpart $t_i^{\mathrm{rev}}$, then embed both with a Conv3D projection $f(\cdot)$

\begin{equation}
x_i = f(t_i),\qquad x_i^{\mathrm{rev}} = f(t_i^{\mathrm{rev}}). 
\end{equation}
Next, we decompose the feature into \emph{appearance} and \emph{motion} subspaces by computing the even and odd components respectively
\begin{equation}
\mathbf{c}_i^{\mathrm{app}} = \tfrac{1}{2}\big(x_i + x_i^{\mathrm{rev}}\big),\qquad
\mathbf{c}_i^{\mathrm{mot}} = \tfrac{1}{2}\big(x_i - x_i^{\mathrm{rev}}\big).    
\end{equation}
We then define the motion score as the feature-norm of the motion component
\begin{equation}
\mathbf{m}_i \;=\; \big\| \mathbf{c}_i^{\mathrm{mot}} \big\|_2
    \;=\; \tfrac{1}{2}\,\big\| x_i - x_i^{\mathrm{rev}} \big\|_2,
\label{eq:motion_score}
\end{equation}
where $\mathbf{m}_i$ is the motion score for tubelet $t_i$. The motion scores of all the tubelets form the motion matrix $\mathcal{M}$, which is computed once before the DTFE, then coupled with the binary keep decision matrix $\mathcal{G}_l$ at each pruning layer $l$. The motion loss is defined as


%
%
\begin{align}
\mathcal{L}_{\text{motion}} = 
-\frac{1}{N_{l}} \sum_{i=1}^{N_{l}} 
(\mathcal{G}_l \cdot \mathcal{M}),
\end{align}
where the negative sign encourages the model to keep tokens with high motion, aligning sparsification with anomaly-relevant dynamics. Finally, the S-WSVAD and P-WSVAL losses are simplified as
\begin{equation}
    \mathcal{L}_{S} = \mathcal{L}_{WSVAD}, \quad \mathcal{L}_{P} = \mathcal{L}_{WSVAD} + \lambda_5 \mathcal{L}_{motion}
\end{equation}


\section{Experiments}
\subsection{Evaluation Metrics}
We follow the standard evaluation procedures established in prior works for temporal anomaly detection evaluation. For spatial localization, we adopt two evaluation protocols. The first is computing the \textit{Temporal IoU (TIoU)} and \textit{Mean IoU (MIoU)} on the anomaly test frames, following prior works \cite{liu2019exploring, stprompt_wu2024weakly, wu2021weakly}. The second is our proposed patch-wise protocol, where we compute the \textit{AUC} and \textit{AP} directly from all patch scores across all test frames. In detail, the frame and patch scores are first computed for each video in the test set. Then, the patch scores are masked by the temporal frame scores. Finally, the patch scores are flattened to compute the \textit{Patch AUC (PAUC)} and \textit{Patch AP (PAP)} metrics. We demonstrate this evaluation protocol in detail in Supp. S3. 

\subsection{Implementation Details}
We adopt a pre-trained VideoMAEv2 \cite{videomaev2}, fine-tuned on Kinetics-710 dataset, to extract the snippet features for the S-WSVAD branch. The WSVAD model applied in both branches is UR-DMU \cite{urdmu_zhou2023dual}. Following standard practices, we compress each video into 200 segments by aggregating and averaging the 16-frame snippet features. The split size $t_z$ in P-WSVAL is set empirically to 2 for all datasets. The learning rate is set to 0.0001 in both S-WSVAD and P-WSVAL. The models are trained with a batch size of 16 for 4000 iterations. The hyperparameters ($\lambda_1$, $\lambda_2$, $\lambda_3$, and $\lambda_4$) are set to their defaults in \cite{urdmu_zhou2023dual} and the motion loss weight $\lambda_5$ is set to 0.01. All experiments are performed under a fixed random seed to ensure reproducibility across runs.

\subsection{Comparison with state-of-the-art }
We first establish a temporal WSVAD baseline (S-WSVAD) by training UR-DMU~\cite{urdmu_zhou2023dual} on VideoMAEv2 snippet features. Compared to the commonly used I3D backbone in prior works~\cite{urdmu_zhou2023dual, rtfm_tian2021weakly, chen2023mgfn, pivad}, VideoMAEv2 provides consistent performance gains, as validated in our experiments. We compare models trained with both feature types to quantify this improvement. Finally, we evaluate our end-to-end SST-WSVADL against prior anomaly localization methods in both temporal detection and spatial localization~\cite{simonyan2014twostream,wang2016TSN,liu2018tc3d,wang2018appearanceARTNet,wang2018non,liu2019exploring,stprompt_wu2024weakly}.
\subsubsection{Temporal Anomaly Detection}
\begin{wraptable}{R}{6.25cm}
\centering
\vspace{-1.2cm}
\begin{minipage}[htb]{\linewidth}
\caption{Comparison with state-of-the-art methods on UCF-Crime. \NA\ indicates metrics not reported or irrelevant. TIoU$_{sub}$ is measured based on the spatial annotations subset provided by \cite{liu2019exploring}.
}
\label{tab:combined_benchmark_compact_ucf}
\resizebox{\linewidth}{!}{%
\begin{tabular}{llcc|cc}
\toprule
 Method & Features &
\cellcolor{Salmon!30}AUC & \cellcolor{Salmon!30}AUC$_{\mathrm{A}}$\cellcolor{Salmon!30}  & \cellcolor{YellowGreen!30}{$\text{TIoU}_{\text{sub}}$}  & 
\cellcolor{YellowGreen!30}{MIoU$_{\text{sub}}$} \\
\midrule
Two Stream~\cite{simonyan2014twostream} & Two-stream
  & 51.20 & \NA & {2.20} & \NA  \\
TSN~\cite{wang2016TSN} & TSN
  & 53.20 & \NA  & {2.60} & \NA  \\
C3D~\cite{tran2015learningc3d} & C3D
  & 70.10 & \NA & {7.20} & \NA   \\
T-C3D~\cite{liu2018tc3d} & C3D
  & 74.50 & \NA   & {10.20} & \NA   \\
ARTNet~\cite{wang2018appearanceARTNet} & ARTNets
  & 75.10 & \NA & {11.40} & \NA    \\
3DResNet~\cite{hara2018canResNet3D} & I3D-ResNet
  & 77.50 & \NA   & {10.30} & \NA  \\
NLN~\cite{wang2018non} & I3D-ResNet
  & 78.90 & \NA & {12.20} & \NA  \\
Liu et al.~\cite{liu2019exploring} & I3D-ResNet
  & 82.00 & \NA & {16.40} & \NA \\
Sultani et al.~\cite{sultani2018real} & I3D  & 67.11 & \NA & {16.82}& \NA  \\
VadCLIP~\cite{wu2024vadclip} & CLIP  & 88.02 & 70.23   & {22.05}& \NA \\
STPrompt~\cite{stprompt_wu2024weakly} & CLIP  & 88.08 & \NA & {23.90}& \NA \\
WSSTAD~\cite{wu2021weakly} & C3D  & 87.65 & \NA & \NA  & 8.98 \\

RTFM \cite{rtfm_tian2021weakly} & I3D  & 84.30 & \NA & \NA    & \NA \\

MGFN \cite{chen2023mgfn} & I3D  & 86.67 & \NA  & \NA & \NA \\

UR-DMU \cite{urdmu_zhou2023dual} & I3D  & 86.97 & 70.81  & \NA & \NA \\

$\pi$-VAD \cite{pivad}& I3D
  &\textbf{ 90.33 }& \textbf{77.77} & \NA  & \NA  \\

\cmidrule{2-6}
\textit{{S-WSVAD \cite{urdmu_zhou2023dual}}} & VideoMAEv2  & 88.07 & 73.98 & \NA & \NA  \\
\cellcolor{extralightgray}\textit{\textbf{SST-WSVADL}} & \cellcolor{extralightgray}VideoMAEv2  & \cellcolor{extralightgray}{88.51} & \cellcolor{extralightgray}{74.35} & \cellcolor{extralightgray}{\textbf{26.25}} & \cellcolor{extralightgray}\textbf{27.92} \\ 
  & 
  & 
  & 
  & \textcolor{red}{(+2.35)}
  & \textcolor{red}{(+18.94)}\\
\bottomrule
\end{tabular}
}
\vspace{-1.0cm}
\end{minipage}
\end{wraptable}
Tables \ref{tab:combined_benchmark_compact_ucf}-\ref{tab:combined_benchmark_compact_xd} present comprehensive temporal and spatial evaluations of the proposed approach on UCF-Crime, MSAD, and XD-Violence datasets. SST-WSVADL delivers competitive joint temporal and spatial performance. In temporal detection, our method improves AUC and AP on MSAD by \textit{+5.49} and \textit{+11.76} respectively over $\pi$-VAD~\cite{pivad}, while achieving competitive performance on UCF-Crime and XD-Violence despite relying on a single modality. Even where SST-WSVADL only matches its temporal-only counterpart on UCF-Crime, grounding detections in localized spatial evidence makes its predictions more interpretable and its background debiasing inspectable. While employing VideoMAEv2 features in S-WSVAD already yields gains over earlier approaches, integrating spatial features through end-to-end optimization in SST-WSVADL further strengthens temporal accuracy. Across both I3D and VideoMAEv2 features (Table \ref{tab:spatial_temporal_block}), our model outperforms its temporal-only counterpart~\cite{urdmu_zhou2023dual} across most configurations, except for I3D on UCF-Crime. Additional class-wise and anomaly-only metrics are provided in Supp. 4.1. These findings indicate that cross-branch information flow not only enables spatial localization but also improves temporal detection. 
\begin{table}
\begin{minipage}[htb]{0.55\linewidth}
\caption{Comparison with state-of-the-art methods on MSAD (temporal detection). \NA\ indicates metrics not reported. }
\label{tab:combined_benchmark_compact_msad}
\resizebox{\linewidth}{!}{%
\begin{tabular}{llcccc}
\toprule
Method & Features &
\cellcolor{Salmon!30}AUC & \cellcolor{Salmon!30}AUC$_{\mathrm{A}}$\cellcolor{Salmon!30} &
\cellcolor{Salmon!30}AP & \cellcolor{Salmon!30}AP$_{\mathrm{A}}$  \\
\midrule
RTFM \cite{rtfm_tian2021weakly} & I3D
  & 86.65 & \NA & \NA & \NA \\

MGFN \cite{chen2023mgfn} & I3D
  & 84.96 & \NA & \NA & \NA \\

TEVAD \cite{chen2023tevad} & I3D
  & 86.82 & \NA & \NA & \NA\\

UR-DMU \cite{urdmu_zhou2023dual} & I3D
  & 85.78 & 67.95 & 67.35 & 75.30  \\

$\pi$-VAD \cite{pivad}& I3D
  & 88.68 & {71.25} & {71.26} & {77.86} \\
\cmidrule{2-6}
\textit{{S-WSVAD \cite{urdmu_zhou2023dual}}} & VideoMAEv2
  & {89.47} & 69.92 & 62.75 & 70.59 \\


\cellcolor{extralightgray}\textit{\textbf{SST-WSVADL }} & \cellcolor{extralightgray}VideoMAEv2
  & \cellcolor{extralightgray}\textbf{94.17} & \cellcolor{extralightgray}\textbf{80.21} & \cellcolor{extralightgray}\textbf{83.02} & \cellcolor{extralightgray}\textbf{85.01} \\

  & 
  & \textcolor{red}{(+5.49)}  & \textcolor{red}{(+8.96)} &  \textcolor{red}{(+11.76)} & \multicolumn{1}{c}{\textcolor{red}{(+7.15)}} \\

\bottomrule
\end{tabular}
} 
\end{minipage}
\hspace{0.1cm}
\begin{minipage}[htb]{0.41\linewidth}
\caption{Comparison with state-of-the-art methods on XD-Violence (temporal detection). \NA\ indicates metrics not reported.}
\label{tab:combined_benchmark_compact_xd}
\resizebox{\linewidth}{!}{%
\begin{tabular}{llcc}
\toprule
Method & Features &
\cellcolor{Salmon!30}AP & \cellcolor{Salmon!30}AP$_{\mathrm{A}}$  \\
\midrule


RTFM \cite{rtfm_tian2021weakly} & I3D & 77.81 & 78.57 \\
UR-DMU \cite{urdmu_zhou2023dual} & I3D & 81.66 & 83.94 \\
TEVAD \cite{chen2023tevad} & I3D& 82.17 & \NA\\
VadCLIP \cite{wu2024vadclip} & CLIP& 84.51 & \NA \\
$\pi$-VAD \cite{pivad} & I3D & {85.37} & 85.79  \\
MissionGNN \cite{yun2025missiongnn} & I3D &\NA & 98.42 \\
\cmidrule{2-4}
\textit{{S-WSVAD \cite{urdmu_zhou2023dual}}} & VideoMAEv2 & {85.33} & {98.69}\\

\cellcolor{extralightgray}\textit{\textbf{SST-WSVADL}} & \cellcolor{extralightgray}VideoMAEv2& \cellcolor{extralightgray}\textbf{86.00}& \cellcolor{extralightgray}\textbf{98.76} \\
 & &\textcolor{red}{(+0.63)} & \multicolumn{1}{c}{\textcolor{red}{(+0.34)}}  \\
\bottomrule
\end{tabular}
} 
\end{minipage}
\end{table}
\begin{table}[htb!]
\centering
\setlength{\tabcolsep}{4pt}
\caption{Performance comparison under different spatio-temporal features. }
\label{tab:spatial_temporal_block}
\resizebox{\linewidth}{!}{%
\begin{tabular}{cccccccccccccc}
\toprule
\multicolumn{1}{c}{Model} & 
\multicolumn{1}{c}{Features} &
\multicolumn{4}{c}{\textbf{UCF-Crime}} &
\multicolumn{4}{c}{\textbf{MSAD}} & 
\multicolumn{4}{c}{\textbf{XD-Violence}}\\
\cmidrule{1-14}

\multicolumn{1}{c}{} & \multicolumn{1}{c}{} & \multicolumn{12}{c}{\cellcolor{Salmon!30}\textit{Temporal Detection}} \\ 
\cmidrule{3-14}
& & \multicolumn{2}{c}{AUC} & \multicolumn{2}{c|}{AUC$_\text{A}$} &  \multicolumn{2}{c}{AUC} & \multicolumn{2}{c|}{AUC$_\text{A}$} &  \multicolumn{2}{c}{AP} & \multicolumn{2}{c}{AP$_\text{A}$}\\ 
\cmidrule{3-14}

\small{\textit{S-WSVAD} \cite{urdmu_zhou2023dual}} & \multirow{2}{*}{I3D}  & \multicolumn{2}{c}{\textit{86.97}} &\multicolumn{2}{c|}{\textit{70.81}} & \multicolumn{2}{c}{\textit{85.78}} & \multicolumn{2}{c|}{\textit{67.95}} & \multicolumn{2}{c}{\textit{81.66}} & \multicolumn{2}{c}{\textit{83.94}} \\ 
\small{SST-WSVADL}  & & \multicolumn{2}{c}{86.50} & \multicolumn{2}{c|}{70.59} & \multicolumn{2}{c}{89.00} & \multicolumn{2}{c|}{70.00} & \multicolumn{2}{c}{82.39} & \multicolumn{2}{c}{98.58} \\ 
\cmidrule(lr){3-14}

 \small{\textit{S-WSVAD \cite{urdmu_zhou2023dual}}} & \multirow{2}{*}{VMAEv2}  & \multicolumn{2}{c}{\textit{88.07}} & \multicolumn{2}{c|}{\textit{73.98}} & \multicolumn{2}{c}{\textit{89.47}} & \multicolumn{2}{c|}{\textit{69.92}} & \multicolumn{2}{c}{\textit{85.33}} & \multicolumn{2}{c}{\textit{98.69}} \\ 
\small{SST-WSVADL} &  & \multicolumn{2}{c}{\textbf{88.51}} & \multicolumn{2}{c|}{\textbf{74.35}} & \multicolumn{2}{c}{\textbf{94.17}} & \multicolumn{2}{c|}{\textbf{80.21}} & \multicolumn{2}{c}{\textbf{86.00}} & \multicolumn{2}{c}{\textbf{98.76}} \\

\multicolumn{1}{c}{} & \multicolumn{1}{c}{} &\multicolumn{12}{c}{\cellcolor{YellowGreen!30}\textit{Spatial Localization}} \\ 

\cmidrule{3-14}

 

&& PAUC & PAP &TIoU& \multicolumn{1}{c|}{MIoU} & PAUC & PAP & TIoU& \multicolumn{1}{c|}{MIoU} & PAUC & PAP & TIoU & MIoU\\ 
\cmidrule{3-14}
 \multirow{2}{*}{\small{SST-WSVADL}} & I3D  
 & 86.01 & 18.46 & 11.69 & \multicolumn{1}{c|}{27.28} 
 & 87.15 & \textbf{40.32} & \textbf{27.48} & \multicolumn{1}{c|}{34.97} 
 &{80.54} & \textbf{72.32} & {19.75} & \textbf{72.50} \\ 
 
& VMAEv2  & 
\textbf{88.45} & \textbf{25.80} & \textbf{14.59}& \multicolumn{1}{c|}{\textbf{29.90}} 
& \textbf{88.40} & 37.46 &  26.27 & \multicolumn{1}{c|}{\textbf{35.02}} 
& \textbf{81.00} & {68.20} & \textbf{29.49} & {63.09} \\ 


\bottomrule
\end{tabular}
}
\end{table}

\subsubsection{Spatial Anomaly Localization} \label{sec:spatial}
Tables \ref{tab:combined_benchmark_compact_ucf} and \ref{tab:spatial_temporal_block}  present the spatial localization results (highlighted in green). The tables report TIoU computed on two evaluation sets: (i) a previously annotated subset of frames by \cite{liu2019exploring} (\textit{TIoU$_{sub}$}), and (ii) the complete set of anomalous frames in the test split, based on our extended annotations. On UCF-Crime, SST-WSVADL with VMAEv2 outperforms both the supervised approach \cite{liu2019exploring} and the VLM-based STPrompt \cite{stprompt_wu2024weakly}, reaching \textit{14.59} TIoU. On MSAD and XD-Violence, our model achieves \textit{26.27} and \textit{29.49} TIoU respectively under the same VMAEv2 configuration. Table \ref{tab:spatial_temporal_block} shows that the choice of backbone affects datasets differently. While VMAEv2 dominates UCF-Crime across all spatial metrics, I3D achieves competitive or superior scores on MSAD and XD-Violence for certain metrics. Due to the absence of publicly available code and trained models for prior localization approaches \cite{landi2019anomaly, stprompt_wu2024weakly}, direct comparison under TIoU, PAUC, and PAP is not feasible.

\begin{figure}[t!]
    \centering
    \includegraphics[width=0.39\linewidth]{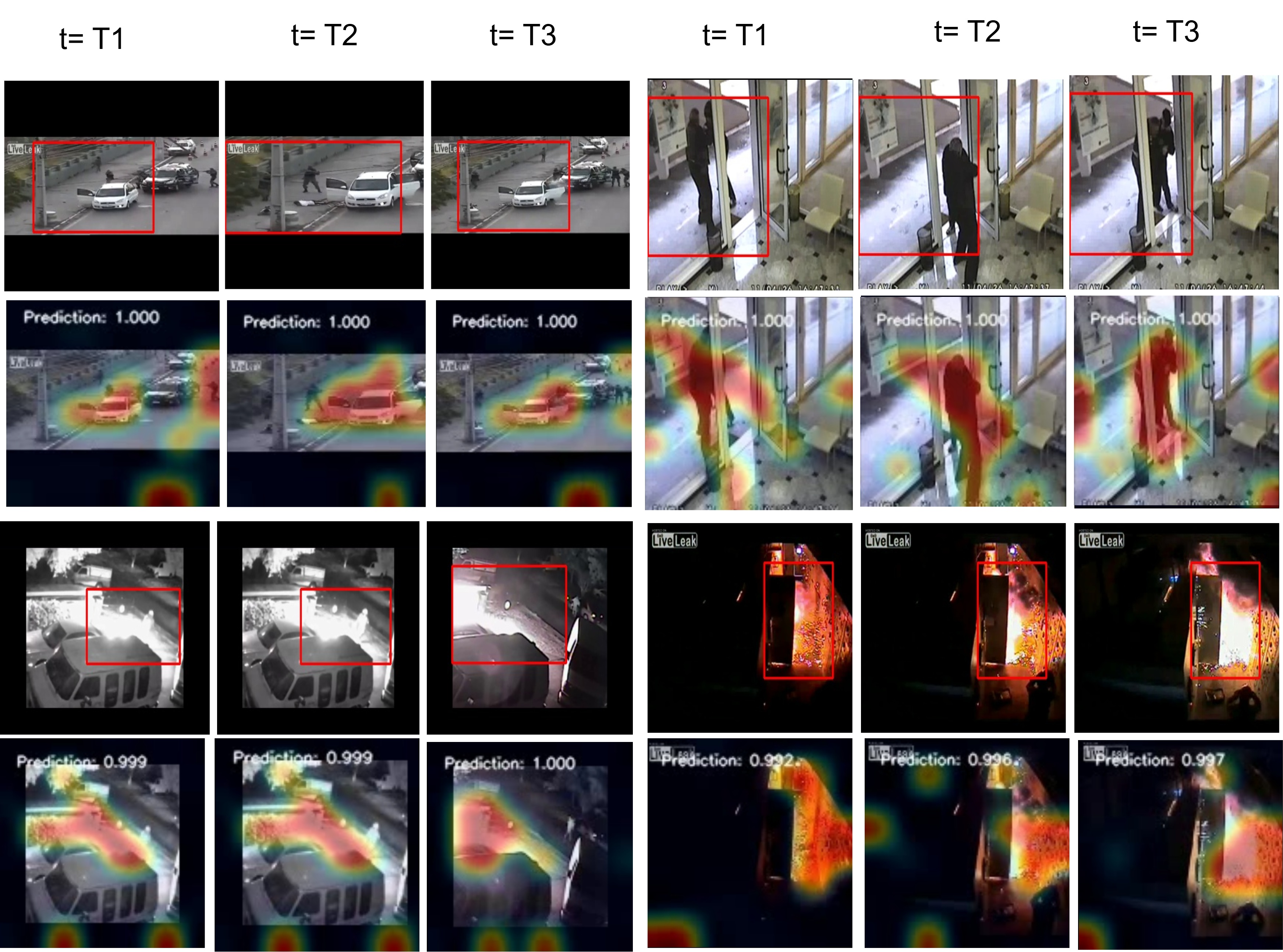}
    \centering
    \includegraphics[width=0.41\linewidth]{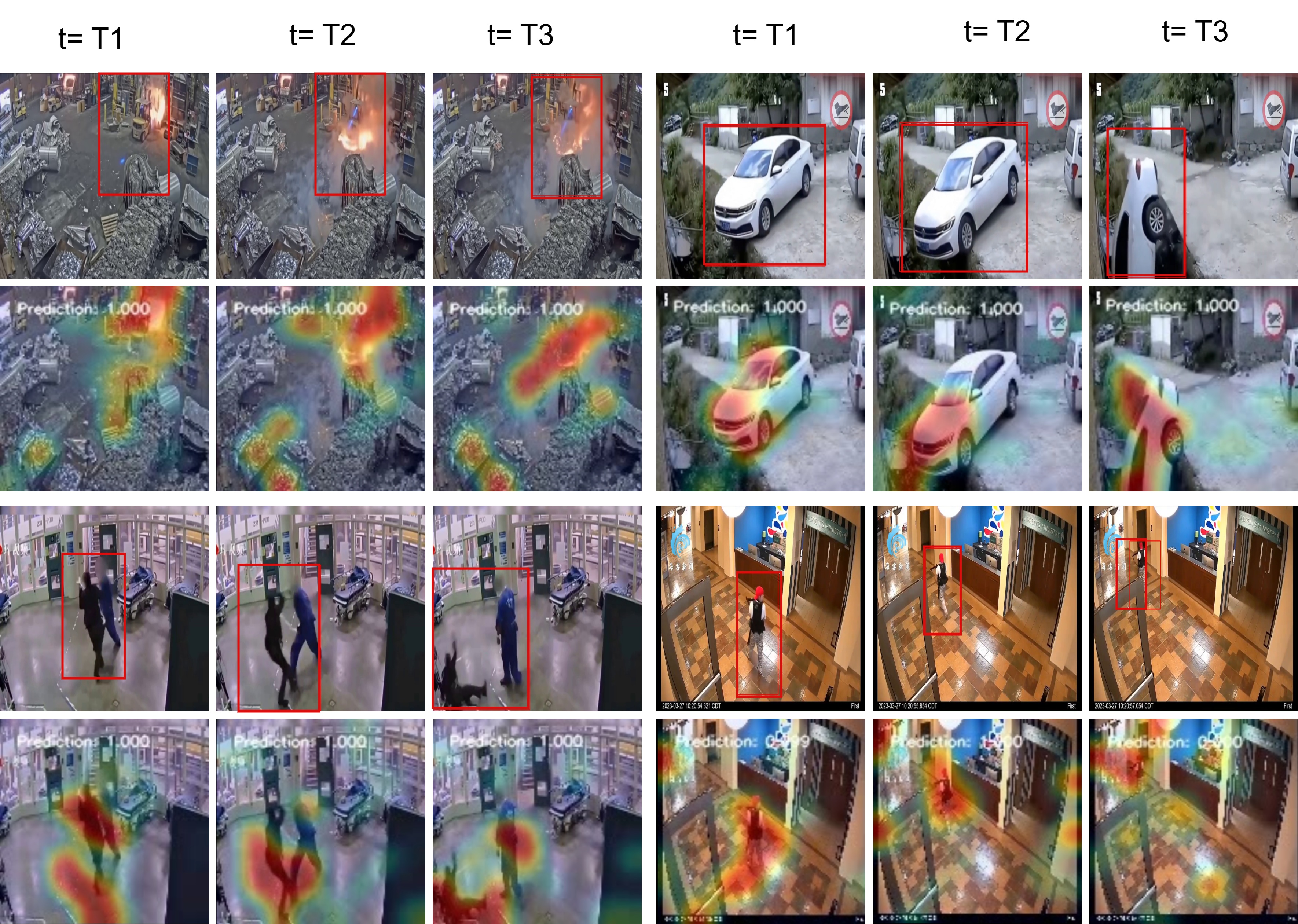}
    \hfill
    \centering
    \includegraphics[width=0.39\linewidth]{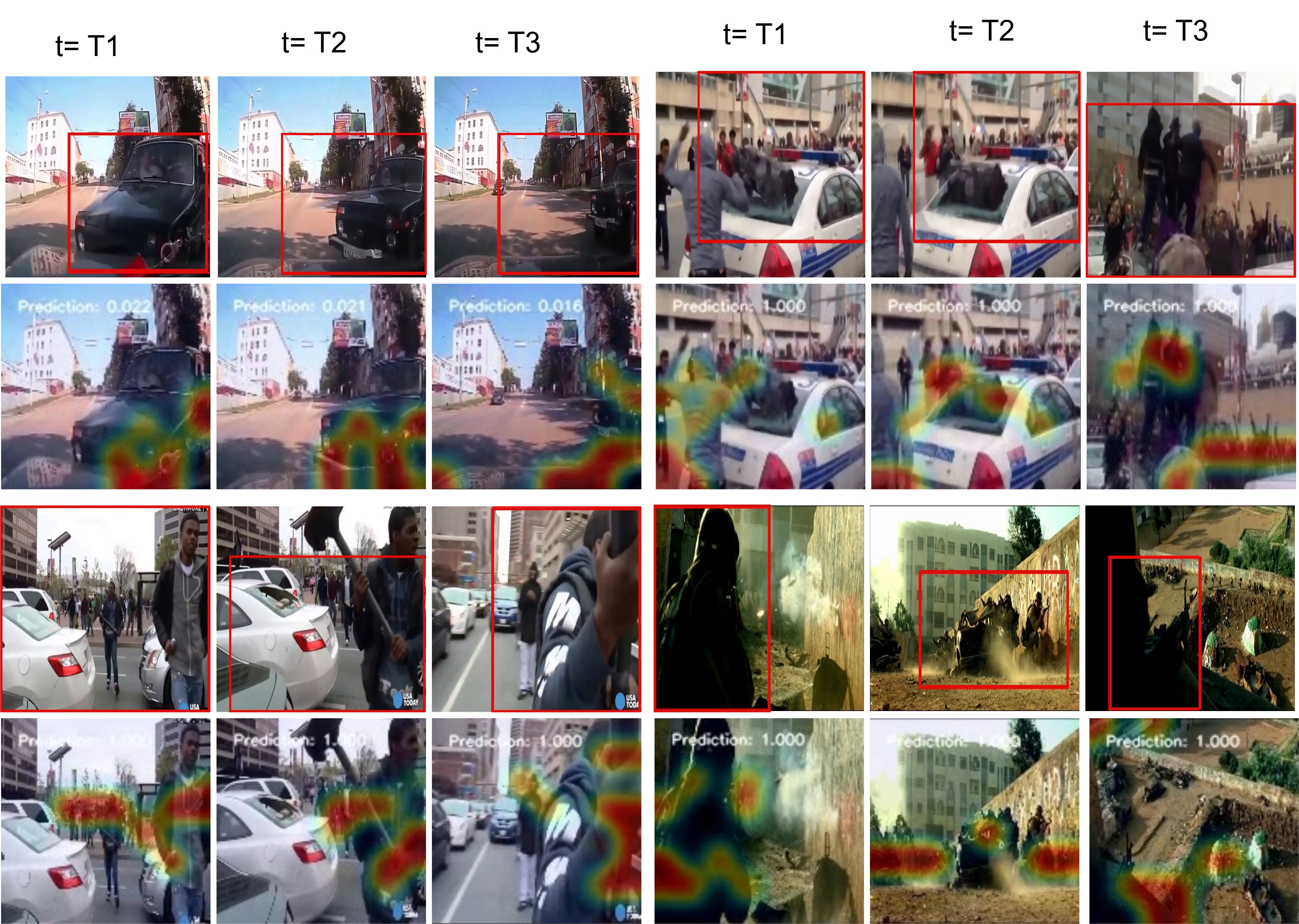}
    \caption{
    Qualitative results (left to right) on UCF-Crime, MSAD, and XD-Violence. Examples are from Arrest024, Arson018, Explosion022, Assault010 (UCF-Crime); Fire-24, TrafficAccident-2, Assault-11, Shooting-4 (MSAD); and v=0qtIjyt-7wg, v=DD3jfKr8e-k, Black.Hawk.Down.2001 (XD-Violence). Zoom in for better visualization. }
    \label{fig:qualitative}
\end{figure}

To assess the performance visually, we produce heatmaps representing the patch outliers. We explain the heatmap generation in detail in Supp. S3.2. Figures \ref{fig:qualitative} and S3 show examples of anomalous frames with localization heatmaps at different timestamps. The visualizations demonstrate the ability of the proposed approach to accurately highlight anomalous regions. In some cases, the heatmaps are more precise than conventional bounding boxes. For instance, the model selectively attends to patches that tightly cover the anomalous actions (e.g., fighting, arrest, traffic accidents), while disregarding irrelevant background areas that are often present within coarse bounding-box annotations. This indicates that SST-WSVADL effectively captures the critical action-relevant cues of anomalous events. Nevertheless, localization remains challenging in some cases due to possible temporal inconsistencies and sensitivity to background motion in normal scenes. We discuss the limitations further in Supp. S6.
%
\subsection{Ablation Studies} 
\paragraph{Effect of Online Clip Proposals.} 
\begin{wraptable}{r}{5.5cm}
\centering
\begin{minipage}[htb!]{\linewidth}
 \vspace*{-1.2cm}
    \caption{Comparison of the effect of proposal selection from the top-k samples on UCF-Crime.}
    \label{tab:topk_ablation}
    {\resizebox{\linewidth}{!}{
    \begin{tabular}{ccc|ccc}
    \toprule
        \multirow{2}{*}{Model} & \multicolumn{2}{c}{\cellcolor{Salmon!30}Temporal}& \multicolumn{3}{c}{\cellcolor{YellowGreen!30}Spatial}   \\
        \cmidrule{2-6}
         & AUC & AUC$_{\text{A}}$ & PAUC  & PAP  & TIoU \\
         \midrule
         Top-1 & \textbf{88.51} & \textbf{74.35} &  88.45 & \textbf{25.80} & 14.59\\
         Rank-2 & 88.29 & 72.46 & \textbf{88.47} & 15.81  & 12.92 \\
         Top-3 & 88.03 & 72.28 &  88.08 & 15.89& \textbf{16.73} \\
         Random(Top-k) & 88.26 & 71.96 & 84.59 & 11.74& 12.55 \\
          Supervised & 87.76 & 73.71 & 80.85 & 8.62 & 11.19 \\
    \bottomrule
    \end{tabular}
    }}
\end{minipage}
\vspace*{-0.7cm}
\end{wraptable} 
We evaluate how different proposal strategies affect performance, including a \textit{supervised} setting and \textit{second-top}, \textit{random-top}, and \textit{multi-top} sampling within the top-\emph{k} predictions. In the supervised setting, P-WSVADL proposals are randomly sampled from ground-truth anomaly snippets without using the temporal branch ranking. As shown in Table~\ref{tab:topk_ablation}, training with top-1 proposals yields the best overall spatio-temporal performance. In contrast, supervised sampling introduces temporal misalignment between branches, preventing cross-attention and degrading performance. Similarly, random or multiple selections from the top-\emph{k} reduce precision by increasing false positives at the patch level, despite comparable TIoU scores. Although top-3 achieves the highest TIoU, top-1 provides better spatial precision and coverage, resulting in the most consistent performance.

\begin{table}[ht]
\centering
\begin{minipage}[ht]{0.45\linewidth}
\caption{Ablation of the SST-WSVADL components on UCF-Crime. } 
\label{tab:ablation-compact}
\resizebox{\linewidth}{!}{
\begin{tabular}{cccc|ccc}
\toprule
\multirow{2}{*}{DTFE} & \multirow{2}{*}{PSA} & \multicolumn{2}{c}{\cellcolor{Salmon!30}Temporal} &\multicolumn{3}{c}{\cellcolor{YellowGreen!30} Spatial} \\
\cmidrule{3-7}
& & AUC &AUC$_{\text{A}}$  & PAUC  & PAP & TIoU \\
\midrule
\xmark & \xmark & 88.00 & 72.68 & 85.68  & 11.75 & 10.79\\
\cmark & \xmark & 87.75 & 73.71 & 86.54 & 12.50 & 12.68 \\
\xmark & \cmark & 88.49 & 74.03 & \textbf{88.45} & 19.23 & 13.98 \\
\midrule
\cmark & \cmark & \textbf{88.51} & \textbf{74.35} & \textbf{88.45}  &  \textbf{25.80}  & \textbf{14.59} \\
\bottomrule
\end{tabular}}
\end{minipage}
\hfill
\begin{minipage}[ht]{0.53\linewidth}
\centering
 \caption{The effect of the motion loss to the P-WSVAL objective on UCF-Crime. }
    \label{tab:motion_loss}
    \resizebox{\linewidth}{!}{
    \begin{tabular}{@{}cccc|ccc@{}}
    \toprule
        \multirow{2}{*}{Model} & \multirow{2}{*}{$\lambda_5$}& \multicolumn{2}{c}{\cellcolor{Salmon!30}Temporal}& \multicolumn{3}{c}{\cellcolor{YellowGreen!30}Spatial}   \\
        \cmidrule{3-7}
         & & AUC & AUC$_{\text{A}}$ & PAUC  & PAP  & TIoU \\
         \midrule
         
      

         w/o. motion loss & \NA &  87.85 & 73.12 & 87.76 & \textbf{33.23} &  12.57 \\
         \midrule
         \multirow{3}{*}{w. motion loss} & 0.01 & \textbf{88.51} & \textbf{74.35} & \textbf{88.45} & 25.80 & \textbf{14.59} \\
         & 0.001 & 88.22 & 73.09 & 88.41 & 25.60 & 13.95 \\
         & 0.0001 & 88.01 & 73.35 & 88.17 & 17.47 & 12.33 \\
    \bottomrule
    \end{tabular}
    }
\end{minipage}
\end{table}

\paragraph{Effect of DTFE and PSA.} Table \ref{tab:ablation-compact} evaluates the individual and joint contributions of the Dynamic Tubelet Feature Encoder (DTFE) and the Patch-Snippet Cross-Attention module (PSA). Removing both yields the weakest spatial performance, indicating that neither dynamic temporal modeling, nor cross-branch interaction alone suffices for reliable localization.
Enabling DTFE alone improves TIoU and PAP, suggesting that dynamic tubelet encoding enhances temporal coherence and produces more discriminative spatio-temporal features. Similarly, introducing PSA alone boosts PAP and PAUC, highlighting the importance of explicit patch-snippet interaction.
Combining both modules consistently improves temporal and spatial metrics, achieving the best overall performance.

\paragraph{Effect of Motion Loss.}
\begin{wraptable}{R}{5.8cm}
\centering
\setlength{\tabcolsep}{3pt}
\vspace*{-0.4cm}
\caption{Impact of motion regularization on sparsified tubelet selection before P-WSVAL on UCF-Crime. MIoU and PAP are evaluated at different retention ratios.}

\label{tab:motion_effect}
\resizebox{\linewidth}{!}{
\begin{tabular}{lccc}
\toprule
Setting & Keep ratios & MIoU (\%) $\uparrow$ & PAP (\%) $\uparrow$ \\
\midrule
\multirow{3}{*}{w/o. motion loss} & 0.5 & 21.78 & 7.10\\
& 0.3 & 34.55 & 10.47 \\
& 0.2 & 10.09 & 4.96 \\
\midrule
\multirow{3}{*}{w. motion loss}  & 0.5 & 17.37 & \textbf{8.84} \\
& 0.3 & \textbf{34.70} & \textbf{14.62} \\
& 0.2 & 6.61 & \textbf{5.21} \\
\bottomrule
\end{tabular}}
\vspace*{-0.85cm}
\end{wraptable} 
Table \ref{tab:motion_loss} shows that incorporating the motion loss consistently improves both temporal and spatial performance. By biasing DTFE toward dynamically informative tubelets, it suppresses static background responses and promotes motion-consistent feature learning. 
This yields stronger temporal discrimination and improved localization overlap (TIoU), at a measurable cost in PAP. The trade-off reflects the pruning behavior analyzed in the shortcut discussion of Table 10: motion regularization prevents the score distribution from concentrating on a small set of high-confidence patches, distributing evidence across genuinely dynamic regions.

\begin{wraptable}{R}{5.8cm}
\vspace*{-1.2cm}
\centering
\setlength{\tabcolsep}{4pt}
\caption{Effect of motion regularization strategy within SST-WSVADL, evaluated on UCF-Crime.}
\label{tab:motion_ablation}
\resizebox{\linewidth}{!}{
\begin{tabular}{@{}lccc@{}}
\toprule
Motion regularization & PAUC & PAP & TIoU \\
\midrule
Variance-based ~\cite{lee2026refinevad}   & \textbf{89.08} & 19.71 & 11.03 \\
Time-reversal (Ours)                & 88.45 & \textbf{25.80} & \textbf{14.59} \\
\bottomrule
\end{tabular}}
\vspace*{-0.8cm}
\end{wraptable}

To investigate how motion regularization shapes the learned sparsification, which is the core mechanism behind the importance region localization stage, we measure the overlap and precision (MIoU and PAP) of the sparsified tubelet set with respect to ground truth. Specifically, we evaluate these metrics at different retention ratios ($\frac{1}{2}$, $\frac{1}{3}$, and $\frac{1}{5}$ of the highest-scoring tubelets) prior to the spatial WSVAD module. This way, we assess the quality of the retained regions before the final anomaly scoring. As shown in Table~\ref{tab:motion_effect}, motion regularization consistently increases PAP across all retention levels, while preserving comparable MIoU at the intermediate sparsity level (0.3). Since the number of retained tubelets is fixed, both settings preserve a similar subset of tokens, stabilizing MIoU. The key difference lies in the composition of the retained set: without motion regularization, additional background tubelets are kept, increasing false positives. Motion regularization suppresses these activations, leading to a higher PAP while maintaining comparable overlap. Qualitative analysis is provided in Supp. S5.1. 

We examine how effective the time-reversal-based motion signal is by comparing it against variance-based motion regularization~\cite{lee2026refinevad} in \Cref{tab:motion_ablation}. Specifically, we substitute our time-reversal signal with MoTAR, which relies on the magnitude of temporal feature change as motion signal. The time reversal regularization instead exploits the directional asymmetry of natural motion, penalizing representations that remain invariant when the temporal order is reversed. Time reversal wins PAP by 6.09 points (relative ~31\%) and TIoU by 3.56 points (relative ~32\%). At comparable ranking performance, time-reversal produces substantially better spatial precision and localization. 

\paragraph{Pruning Behavior} \label{supp:sparsification} 

We analyse the learned pruning rates per layer under different training settings in Table S7. As observed, under \textit{unguided pruning}, the model drops too many tokens than necessary to adequately represent anomalous regions. To validate this, we compare the anomaly patch coverage (Table S8) with the amount of retained tokens according to the learned pruning rates (Table S7). In UCF-Crime, roughly 25\% of anomalous frames are covered by 20 out of 64 patches, suggesting that the final layer should retain at least 30\% of tubelets (e.g., 38 of 128 when $t_z$=2). However, as shown in the first row of Table S7, the model prunes far more aggressively, possibly learning a shortcut rather than identifying meaningful tubelets for the anomaly localization task.
\begin{wraptable}{R}{6cm}
\vspace*{-1.2cm}
\centering
\setlength{\tabcolsep}{3pt}
\caption{Hard vs.\ soft pruning under SST-WSVADL and motion loss.}
\label{tab:soft_vs_hard_pruning}
\resizebox{\linewidth}{!}{
\begin{tabular}{lcc|ccc}
\toprule
\multirow{2}{*}{Pruning} & \multicolumn{2}{c}{\cellcolor{Salmon!30}Temporal}& \multicolumn{3}{c}{\cellcolor{YellowGreen!30}Spatial}   \\
 
 & AUC & AP & PAUC & PAP & TIoU\\
\midrule
Soft~\cite{softpruning_kong2022spvit}     & 88.37 & \textbf{42.80} & 87.65 & 18.69 & 14.48 \\
Hard~\cite{rao2021dynamicvit}     & \textbf{88.52} & 37.05 & \textbf{88.45} & \textbf{25.80} & \textbf{14.59} \\
\bottomrule
\end{tabular}}
\vspace*{-0.6cm}
\end{wraptable}

Another shortcut signature is observed when we swap the adopted hard pruning in SST-WSVADL with soft pruning~\cite{softpruning_kong2022spvit} (\Cref{tab:soft_vs_hard_pruning}). Unlike dropping tubelets discretely with a binary decision, soft pruning scores are turned into continuous mask via a $Sigmoid(.)$ function, making them differentiable. This leaves the patch-level score distribution diffuse, where no token is forced to commit either way. The score dispersion affects the per-patch discriminativeness (PAUC and PAP) more than frame-level ranking (AUC and AP). As AP actually \emph{improves} under soft pruning ($37.05 \rightarrow 42.80$) while PAP drops sharply ($25.80 \rightarrow 18.69$), the model becomes better at ranking \emph{which} frames are anomalous without needing to know \emph{where} the anomaly is within frames. Hard pruning eliminates this route, forcing tokens to commit spatially or be discarded entirely. This trades a small AP cost for substantially sharper patch-level grounding. 


\paragraph{Computational Complexity.} We report the computational complexity analysis in Supp. S4.2. 


\subsection{Scene-Bias Analysis}
To assess how much scene context influences SST-WSVADL scoring versus the S-WSVAD base, we adopt the \emph{Bias-AUC} audit of~\cite{vad_audit}, computed on normal frames using the authors' UCF-Crime scene-factor annotations. Bias-AUC equals $0.5$ when a model is insensitive to a scene group; larger deviations indicate scene-conditional scoring.
\begin{wraptable}{R}{6cm}
\vspace*{-1.2cm}
\centering
\setlength{\tabcolsep}{5pt}
\caption{Scene Bias-AUC across scene factors~\cite{vad_audit}, split by backbone. } \label{tab:scene_bias}
\resizebox{\linewidth}{!}{
\begin{tabular}{lcc}
\toprule

Model & features & Mean $|B-0.5| (\downarrow)$ \\
\midrule
S-WSVAD (base) \cite{urdmu_zhou2023dual}   & VideoMAEv2 & 0.113 \\
\textbf{SST-WSVADL (ours)}             & VideoMAEv2   & {0.100} \\
Universal MIL~\cite{sultani2018real} & VideoMAEv2 & 0.086 \\
PiVAD \cite{pivad}                      & VideoMAEv2   & \textbf{0.084}  \\

\midrule
TEVAD \cite{chen2023tevad}             & I3D    & 0.119\\
Universal MIL~\cite{sultani2018real} & I3D & 0.096 \\
S-WSVAD (base) \cite{urdmu_zhou2023dual}& I3D   & 0.093 \\
\textbf{SST-WSVADL (ours)}             & I3D   & {0.080} \\
PiVAD \cite{pivad}                      & I3D   & \textbf{0.064}  \\
\midrule

VadCLIP \cite{wu2024vadclip} & CLIP   & 0.112  \\




\bottomrule
\end{tabular}}
\vspace*{-0.6cm}
\end{wraptable}


Table~\ref{tab:scene_bias} reports mean $|\text{Bias-AUC}-0.5|$ averaged across scene factors, split by backbone. Since S-WSVAD reuses UR-DMU~\cite{urdmu_zhou2023dual}, it serves as the spatial-free configuration. Under a fixed backbone, adding the sparse spatial branch reduces mean scene bias by $0.013$ under both VideoMAEv2 and I3D. The backbone itself has a larger effect than these components. Switching from VideoMAEv2 to I3D reduces mean bias by $0.020$ for both UR-DMU and SST-WSVADL. Spatial localization also remains modest in absolute terms, so SST-WSVADL is best understood as a transparent baseline rather than a solution to scene bias. Still, the consistent $0.013$ reduction under fixed-backbone comparison shows that scene bias in a WSVAD algorithm is measurably reducible under spatial conditioning.


\section{Conclusion} In this work, we released a public benchmark for video anomaly localization across three datasets: UCF-Crime, XD-Violence, and MSAD. Coupled with a method-agnostic evaluation protocol, it enables interpretability-oriented assessment of whether anomaly predictions are grounded in authentic event regions rather than background context. We further proposed SST-WSVADL, a sparse spatio-temporal framework that unifies snippet-level temporal reasoning with patch-level localization. Through motion-aware sparsification during training, SST-WSVADL steers feature learning away from background-dominant regions toward authentic anomaly cues. Without relying on detectors, dense annotations, or vision-language models, it delivers competitive performance across all benchmarks, while achieving measurable but modest reductions in scene bias. Together, these contributions support progress toward more transparent and accountable weakly-supervised anomaly detection. In future work, we plan to extend this sparse formulation to multimodal scenarios and explore different sparsification strategies and granularity levels.

\putbib[main]
\end{bibunit}


\end{document}